\documentclass[nohyperref]{article}

\usepackage{microtype}
\usepackage{graphicx}
\usepackage{subfigure}
\usepackage{booktabs} % for professional tables

\usepackage{hyperref}

\usepackage[accepted]{icml2024}

\usepackage{amsmath}
\usepackage{amssymb}
\usepackage{mathtools}
\usepackage{amsthm}
\usepackage{enumitem}
\usepackage{float}
 \usepackage{algorithm}
\usepackage[noend]{algpseudocode}
\usepackage[linesnumbered,ruled,vlined,algo2e]{algorithm2e}
\usepackage[capitalize,noabbrev]{cleveref}

\theoremstyle{plain}

\theoremstyle{definition}

\theoremstyle{remark}

\usepackage[textsize=tiny]{todonotes}

\icmltitlerunning{Adsorbate placement via conditional denoising diffusion}

\begin{document}

\twocolumn[
\icmltitle{AdsorbDiff: Adsorbate Placement via Conditional Denoising Diffusion }



\icmlsetsymbol{equal}{*}

\begin{icmlauthorlist}
\icmlauthor{Adeesh Kolluru}{cmu}
\icmlauthor{John R. Kitchin}{cmu}

\end{icmlauthorlist}

\icmlaffiliation{cmu}{Department of Chemical Engineering, Carnegie Mellon University}

\icmlcorrespondingauthor{Adeesh Kolluru}{akolluru@andrew.cmu.edu}
\icmlcorrespondingauthor{John R. Kitchin}{jkitchin@andrew.cmu.edu}
\icmlkeywords{Machine Learning, ICML}

\vskip 0.3in
]



\printAffiliationsAndNotice{} % otherwise use the standard text.

\begin{abstract}
Determining the optimal configuration of adsorbates on a slab (adslab) is pivotal in the exploration of novel catalysts across diverse applications. Traditionally, the quest for the lowest energy adslab configuration involves placing the adsorbate onto the slab followed by an optimization process. Prior methodologies have relied on heuristics, problem-specific intuitions, or brute-force approaches to guide adsorbate placement. In this work, we propose a novel framework for adsorbate placement using denoising diffusion. The model is designed to predict the optimal adsorbate site and orientation corresponding to the lowest energy configuration. Further, we have an end-to-end evaluation framework where diffusion-predicted adslab configuration is optimized with a pretrained machine learning force field and finally evaluated with Density Functional Theory (DFT). Our findings demonstrate an acceleration of up to 5x or 3.5x improvement in accuracy compared to the previous best approach. Given the novelty of this framework and application, we provide insights into the impact of pre-training, model architectures, and conduct extensive experiments to underscore the significance of this approach.

\end{abstract}

\section{Introduction}

% Problem motivation and general intro
Heterogenous catalysis plays an important role in developing chemicals in industries, environmental protection through converters, and the synthesis of alternative fuels \cite{liu2017heterogeneous, zitnick2020introduction}. Modeling these chemical reactions involve an intermediate adsorbate on a catalyst slab which determines the efficacy of the catalyst for that particular reaction. Discovering a novel catalyst computationally involves screening through billions of candidates and finding the lowest energy configuration. 

% heuristic methods
Finding the lowest energy configuration for an adsorbate and slab requires a global optimum (which is non-convex) search across different sites on the slab. Conventional approaches solve this in two steps - (1) heuristically place the adsorbate on certain important sites and (2) perform optimization with quantum mechanical calculators like Density Functional Theory (DFT) on each of these sites. The lowest energy site out of these is considered for calculating adsorption energy, which is a thermodynamic descriptor for how good that catalyst is. With recent advances in machine learning methods for predicting forces, it has become possible to perform optimization with ML force fields (MLFFs) instead of Density Functional Theory (DFT) making this process faster and easier to test many sites and find better minima. These ML force fields are trained on DFT data to predict energies and forces corresponding to different adslab configurations.

% Talk more about recent approaches for solving this. AdsorbML and DeepMind paper

The recent release of the OC20-Dense dataset \cite{lan2023adsorbml} signifies a significant advancement in the computation of the lowest energy adslab configuration. This work employs a blend of heuristic and random adsorbate placements across 100 sites, with subsequent optimizations across each site using Density Functional Theory (DFT) to calculate adsorption energy. The study further introduces AdsorbML, a paradigm characterized by a brute-force exploration of initial adsorbate placements. Employing pre-trained machine learning (ML) force fields from OC20, AdsorbML streamlines the optimization process, culminating in the determination of the lowest energy adsorbate-slab (adslab) configuration. The predictive accuracy of these configurations is rigorously validated against DFT single-points or complete DFT optimization. This hybrid approach results in a computational acceleration of 2000-fold in adsorption energy calculations compared to the sole reliance on DFT calculations.

Recent developments in graph neural network (GNN) based ML architectures have increased the accuracies of adsorption energy prediction significantly by encoding geometric information of atoms in more explicit ways. However, there's little to no work done on improving the adsorption site prediction which could help us get away with the currently used brute-force approach. 

In this work, we develop a novel conditional denoising diffusion framework for adsorbate placement. We first formulate a diffusion framework over the space of the 2D translation and 3D rigid rotation of an adsorbate molecule over the slab considering periodic boundary conditions (PBC) of the slab. Through the learned diffusion process, we sample the most stable site by iteratively updating the center of mass of adsorbate and rigid orientation. Performing a naive unconditional diffusion framework on the most optimal adsorbate site and orientation --- corresponding to the lowest energy adslab configuration out of 100 densely sampled calculations in OC20-Dense --- leads to throwing away  99\% of DFT optimal energy data. Therefore, we modify the diffusion training to be conditional on relative energies (relative across densely sampled sites of an adslab combination). This leads to significant improvements in accuracies and sample efficiency during diffusion training. After sampling for the optimal site and orientation of adsorbate on the slab, we perform ML force field (MLFF) optimization and DFT single-point verification similar to AdsorbML. This comprehensive end-to-end evaluation helps in robust assessment of the practical impact of the learned diffusion model. 

There have been significant advances in diffusion generative models in molecular and material discovery, and analogous problems in molecular docking on proteins. However, this is the first work to frame the adsorbate placement problem considering all its symmetries with the slab in a diffusion framework. Intuitively, the reverse diffusion process of AdsorbDiff helps in skipping multiple minima sites due to its energy-based conditional sampling which is followed by a local optimization with a DFT-learned MLFF to find a global optimum. To facilitate further research on this problem, we provide comprehensive results on the importance of GNN architectures for the diffusion task, show the importance of pretraining, and demonstrate the success of our approach to in-distribution (ID) and out-of-distribution (OOD) splits. 

The summary of contributions of this work are - 
\begin{itemize}[label=\textbullet, itemsep=0pt, parsep=0pt,topsep=0pt]

\item  We propose AdsorbDiff, a novel conditional denoising diffusion framework designed to leverage the translation, rotation, and periodic symmetries inherent in adsorbate and slab interactions. Additionally, this framework is adept at efficiently predicting the lowest energy site by conditional training on relative energies.

\item We present our results in a comprehensive end-to-end evaluation framework, integrated with DFT, to accurately gauge the true capability of our approach in predicting optimal adsorption energies.

\item We achieve a 31.8\% success rate, 3.5x higher than the naive AdsorbML baseline of 9.1\% with a single site prediction. Alternatively, we demonstrate that a comparable level of accuracy could be achieved by AdsorbML by employing 5x more placements.

\item We demonstrate that pretraining on large-scale local optimization data can significantly improve the results on the search for global optima.

\item We show that diffusion results exhibit insignificant dependence on GNN architectures, in contrast to the notable differences observed for the same architectures when trained on DFT forces.

\item We highlight the model's generalization capabilities to previously unseen adsorbates and slabs.

\end{itemize}

\begin{figure*}[t]
    \centering
    \includegraphics[width=\textwidth]{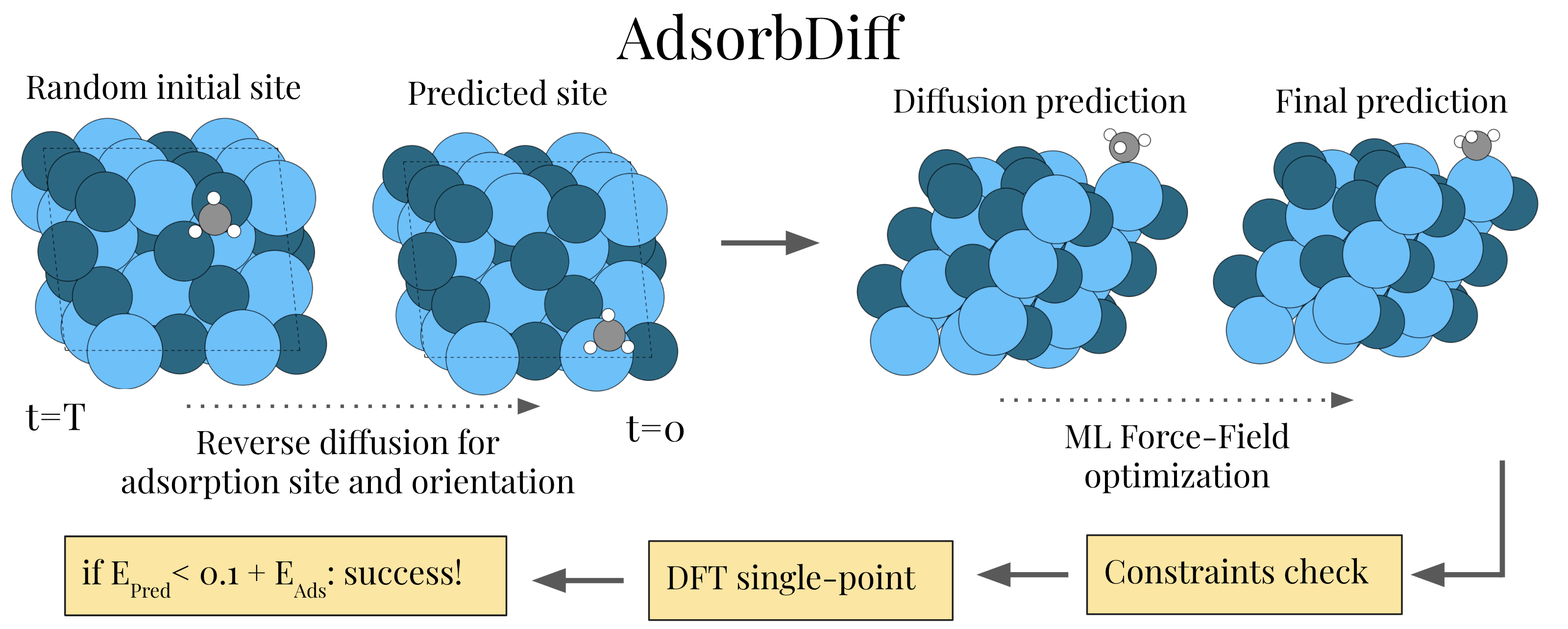}
    \caption{Overview of AdsorbDiff: Random initial site and orientation for the adsorbate are selected, followed by sampling over 2D translation, 3D rigid rotations, and considering periodic boundary conditions (PBC) to predict the optimal site and orientation. MLFF optimization is then conducted from the predicted site with a fixed interstitial gap until convergence. The final prediction undergoes constraint verification, and DFT verification is performed on valid structures to calculate success rates.}
    \label{fig:adsorbdiff_main}
\end{figure*}

\section{Background and Related Work}

\textbf{Force-fields:}
% Force field applications in performing MD and structure optimization
Energy and forces (as a gradient of energy with respect to positions) are calculated using ab initio quantum mechanical methods like Density Functional Theory (DFT). ML models can be trained to predict these energies and forces, and are called ML force-fields (MLFFs). These force fields can be utilized to perform structure optimization to get the lowest energy structures. 

\textbf{Optimization:}
% explain the optimization problem and 
For adsorption energy prediction, we start with an optimized adsorbate and slab, place the adsorbate on a slab, and perform optimization to get an adslab configuration with the lowest energy. Usually, second-order optimizers like BFGS, L-BFGS, Conjugate gradient descent, etc are used to solve this optimization problem. Since this is non-convex, the initial guess of adsorbate placement or the strategy of optimization is critical to finding an adslab configuration corresponding to the global optimum. AdsorbML \cite{lan2023adsorbml} method starts with combining heuristic and random initial placements which is a brute-force approach to finding better minima. "Easy Potential" from \cite{schaarschmidt2022learned} trains a simple harmonic potential to guess this initial placement. Learn2Hop \cite{merchant2021learn2hop} also learns the optimization landscape to navigate through better and hop through local minima. There are approaches like minima hopping that help in navigating through the entire optimization landscape with a force-field \cite{jung2023machine} and help in finding better minima, but these could be computationally expensive. 

\textbf{GNNs:}
% model architectures
Message-Passing Neural Networks (MPNN) are a class of graph neural networks (GNN) that are utilized across material property prediction tasks. Different architectures encode the geometric information in different ways. SchNet \cite{schutt2018schnet} only encodes the distance information. Including more explicit geometric features have improved the model prediction as DimeNet \cite{gasteiger2020directional, gasteiger2020fast} incorporates triplets. SphereNet \cite{liu2021spherical}, GemNet \cite{gasteiger2021gemnet, gasteiger2022gemnet} incorporates complete geometric information explicitly by giving triplets and quadruplets information. PaiNN \cite{schutt2021equivariant} incorporates directional information and applies only linear operations on those features. Equivariant models like NequIP \cite{batzner20223}, Allegro \cite{musaelian2023learning}, MACE \cite{batatia2022mace}, SCN \cite{zitnick2022spherical}, Equiformer \cite{liao2022equiformer, liao2023equiformerv2} utilize spherical harmonics in representing the geometric features. 

% Force predictions can either be done by taking a gradient on energy or by predicting a vector prediction directly from the model. Methods using the former approach give energy-conserving predictions but the latter approaches can be useful for large models and in computational cost reduction \cite{kolluru2022open}. 

\textbf{Diffusion Models:}
%general diffusion models
Diffusion models are a class of generative models that have shown impressive results across different domains starting from computer vision \cite{dhariwal2021diffusion, croitoru2023diffusion}, language models \cite{gong2022diffuseq}, temporal data modeling, to applications in molecules \cite{xu2022geodiff, xu2023geometric, arts2023two, hoogeboom2022equivariant, jing2022torsional}, proteins \cite{wu2022protein, trippe2022diffusion, watson2022broadly, watson2023novo} and materials \cite{xie2021crystal, fu2023mofdiff, zeni2023mattergen, merchant2023scaling, yang2023scalable}.

There are different kinds of formulations proposed for diffusion models like denoising diffusion probabilistic models (DDPMs), score-based generative models (SGMs), and stochastic differential equations (Score SDEs) \cite{yang2023diffusion}. Many of these formulations have been adapted to problems in molecular and material discovery. For example, CDVAE \cite{xie2021crystal} adapts concepts from noise-conditioned score networks (NCSN) for bulk discovery. Conditional diffusion has also been recently utilized across proteins \cite{krishna2024generalized}, catalyst and materials \cite{zheng2023towards} for generating structures with required properties.

Diffusion models have also been recently utilized for molecular docking on proteins \cite{corso2022diffdock}. Although this problem is somewhat analogous to placing adsorbate on a slab, as far as we know there hasn't been previous work on formulating adsorbate placement in a diffusion framework. AdsorbDiff also differs from molecular docking in several key aspects -- 2D translation formulation, periodic boundary conditions, conditional denoising formulation, and the requirement of DFT level accuracy as opposed to simple force-fields for proteins making our end-to-end evaluation with DFT critical.

\section{AdsorbDiff}

\subsection{Overview}
The objective of this research is to enhance the efficiency of adsorption energy calculation, representing the lowest energy configuration of an adsorbate on a slab. The methodology of this work involves the initial placement of an adsorbate on a random site within the 2D surface of the slab, followed by reverse diffusion to predict the optimal adsorption site and orientation. Employing machine learning force field optimization, the structure undergoes iterative updates with an optimizer until forces converge close to 0. Subsequently, the final structure is verified for compliance with constraints essential for defining adsorption energy. On the optimized structure, a single Density Functional Theory (DFT) calculation is conducted to obtain the predicted energy ($E_{Pred}$). A successful outcome is determined by the predicted energy being within 0.1 eV or lower than the DFT baseline of adsorption energy in OC20-Dense data, indicating the model's ability to provide a comparable or superior estimate of adsorption energy (shown in Figure \ref{fig:adsorbdiff_main}). The code is open-sourced with MIT License\footnote{\url{https://github.com/AdeeshKolluru/AdsorbDiff}}.
  
\subsection{Adsorbate placement}

Various adsorbate placement strategies were explored for the OC20-Dense dataset, incorporating a combination of heuristic and random approaches. Specifically, 100 sites were selected for each adslab configuration, utilizing a blend of heuristic and random placements. The heuristic placement involved strategically situating the adsorbate's binding site on either an on-top site, hollow site, or bridge site, with a specified interstitial gap denoting the distance between the connecting atom of the slab and the corresponding adsorbate atom. Additional random sites are introduced through the random rotation of the adsorbate along the normal of the slab, accompanied by a slight translational wobble along the surface from the heuristic site.

\subsection{Diffusion for adsorbate placement}
\label{sec:diff_method}

In this work, our objective is to develop a diffusion model aimed at predicting the adsorbate orientation and site corresponding to the lowest energy, as established through benchmarking with the OC20-Dense dataset. 

The adsorbate motion is constrained within a manifold ($M_c$) and utilizes the combined action group ($A$), as described in DiffDock \cite{corso2022diffdock}. This manifold permits the adsorbate to navigate towards configurations with low-energy adslab states through a combination of translations, rotations, and torsion angle adjustments. Note, for fair comparisons with our baselines, torsion angle alterations are disregarded in our analysis due to the smaller size of the adsorbate employed in this study. This approach aligns with the methodology of AdsorbML, which does not introduce randomness in torsion angles as part of its benchmark.

In our framework, we specifically consider translations in the 2D plane parallel to the slab while accounting for periodic boundary conditions (PBC). The z-coordinate is meticulously aligned to denote the normal direction of the slab and the diffusion process is executed across the xy-coordinates. Therefore, the adsorbate movements are associated with the 2D translation group $T(2)$, and rigid rotations are modeled using the $SO(3)$ group. The translation operation, denoted as $A_{\text{tr}} : T(2) \times \mathbb{R}^{2n} \to \mathbb{R}^{2n}$, is defined as $A_{\text{tr}}(r, x)_i = x_i + r$, employing the isomorphism $T(2) \cong \mathbb{R}^2$, where $x_i \in \mathbb{R}^2$ represents the position of the $i$-th adsorbate atom. Similarly, the rotation operation, denoted as $A_{\text{rot}} : \text{SO}(3) \times \mathbb{R}^{3n} \to \mathbb{R}^{3n}$, is defined by $A_{\text{rot}}(R, x)_i = R(x_i - \bar{x}) + \bar{x}$, where $\bar{x} = \frac{1}{n} \sum_{i} x_i$, signifying rotations around the center-of-mass of the adsorbate.

For the initial coordinates of adsorbate, we select a random point on the slab. This point is considered as the center-of-mass of the adsorbate in fractional coordinates. We then convert from fractional coordinates to real coordinates and perform a reverse diffusion process to get to the lowest energy site (as shown in Algorithm 1).

\begin{figure}[h]
    \centering
    \includegraphics[width=\textwidth/2]{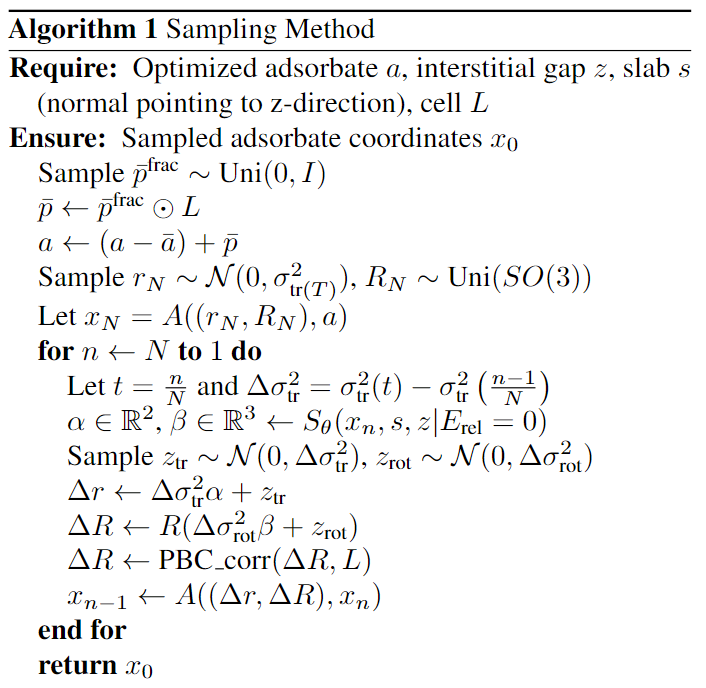}

\end{figure}
\begin{figure}[h]
    \centering
    \includegraphics[width=\textwidth/2]{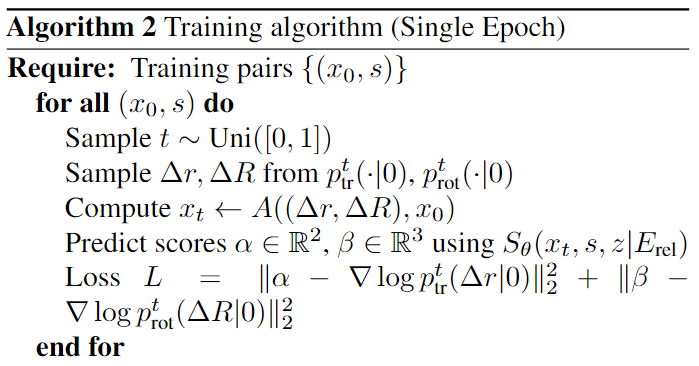}

\end{figure}

% \begin{algorithm}[h]
% \caption{Sampling Method}
% \label{alg:samping}
%     \begin{algorithmic}
%      \Require Optimized adsorbate $a$, interstitial gap $z$, slab $s$ (normal pointing to z-direction), cell $L$ 
%     \Ensure Sampled adsorbate coordinates $x_0$
%     \STATE Sample $\bar{p}^{\text{frac}} \sim \text{Uni}(0, I)$
%     \STATE $\bar{p} \gets \bar{p}^{\text{frac}}\odot L$ %
%     \STATE $a \gets (a - \bar{a}) + \bar{p}$
%     \STATE Sample $r_N \sim \mathcal{N}(0, \sigma_{\text{tr}(T)}^2)$, $R_N \sim \text{Uni}(SO(3))$
%     \STATE Let $x_N = A((r_N, R_N), a)$
    
%         \FOR{$n \gets N$ \textbf{to} $1$}
%             \STATE Let $t = \frac{n}{N}$ and $\Delta \sigma_{\text{tr}}^2 = \sigma_{\text{tr}}^2(t) - \sigma_{\text{tr}}^2\left(\frac{n-1}{N}\right)$
%             \STATE $\alpha \in \mathbb{R}^2$, $\beta \in \mathbb{R}^3 \gets S_{\theta}(x_n, s, z | E_\text{rel}=0)$
%             \STATE Sample  $z_{\text{tr}} \sim \mathcal{N}(0, \Delta \sigma_{\text{tr}}^2)$, $z_{\text{rot}} \sim \mathcal{N}(0, \Delta \sigma_{\text{rot}}^2)$
%             \STATE $\Delta r \gets \Delta \sigma_{\text{tr}}^2 \alpha + z_{\text{tr}}$
%             \STATE $\Delta R \gets R(\Delta \sigma_{\text{rot}}^2 \beta + z_{\text{rot}})$
%             \STATE $\Delta R \gets \text{PBC\_corr}(\Delta R, L)$
%             \STATE $x_{n-1} \gets A((\Delta r, \Delta R), x_n)$
%         \ENDFOR
    
%     \Return $x_0$
    
%     \end{algorithmic}
% \end{algorithm}

The work conducted by De et al. \cite{de2022riemannian} and Corso et al. \cite{corso2022diffdock} has demonstrated the applicability of the diffusion framework to Riemannian manifolds. In this context, the score model constitutes the tangent space, and a geodesic random walk serves as the reverse stochastic differential equation (SDE) solver. The score model is trained using denoising score matching \cite{song2019generative}, wherein a score function $s_{\theta}(x)$ is learned to approximate the gradient of the probability density ${\nabla}_xp(x)$ at varying noise levels (as shown in Algorithm 2).

% \begin{algorithm}[H]
% \caption{Training algorithm (Single Epoch)}
% \label{alg:training}
% \begin{algorithmic}
% \REQUIRE Training pairs $\{(x_0, s)\}$

% \FORALL{$(x_0, s)$}
%     \STATE Sample $t \sim \text{Uni}([0, 1])$
%     \STATE Sample $\Delta r, \Delta R$ from $p_{\text{tr}}^t(\cdot | 0)$, $p_{\text{rot}}^t(\cdot | 0)$
%     \STATE Compute $x_t \gets A((\Delta r, \Delta R), x_0)$
%     \STATE Predict scores $\alpha \in \mathbb{R}^2$, $\beta \in \mathbb{R}^3$ using $S_{\theta}(x_t, s, z | E_\text{rel})$
%     % \STATE Loss
%     \STATE Loss $L = \|\alpha - \nabla \log p_{\text{tr}}^t(\Delta r | 0)\|_2^2 + \|\beta - \nabla \log p_{\text{rot}}^t(\Delta R | 0)\|_2^2$
% \ENDFOR

% \end{algorithmic}
% \end{algorithm}

The learned scores for translations and rotations are treated as independent entities, assuming the tangent space is a direct sum of individual tangent spaces, with contributions from torsion being neglected. The forward SDE for both translation and rotation is defined as $dx = \sqrt{\frac{d\sigma^2(t)}{dt}dw}$, where $w$ represents the corresponding Wiener process. In the translational scenario within $T(2)$, the model learns a score for a standard Gaussian distribution with variance $\sigma^2(t)$. For rotations in $SO(3)$, the diffusion kernel is governed by the $IGSO(3)$ distribution, which can be sampled in the axis-angle parameterization. This involves sampling a unit vector $\omega' \in so(3)$ uniformly and a random angle $\omega$ from the interval $[0, \pi]$, as outlined by Equations \ref{eq:igso3_dist_1} and \ref{eq:igso3_dist_2}. The score of diffusion kernel is defined in Equation \ref{eq:score_diffkernel}. The computation of $R' = R(\omega \hat{\omega})R$, where $R$ is the result of applying the Euler vector $\omega \hat{\omega}$ to $R$, has been established in prior work by Yim et al. \cite{yim2023se}. To efficiently carry out the score computation and sampling processes, it is feasible to precompute the truncated infinite series and interpolate the cumulative distribution function (CDF) of $p(\omega)$.

\begin{gather}
\label{eq:igso3_dist_1}
    p(\omega) = \frac {1 - \cos (\omega)}{\pi} f(\omega) \\
\label{eq:igso3_dist_2}
    \begin{aligned}
    f(\omega) &= \sum_{l=0}^{\infty} (2l + 1) \exp\left(-\frac{l(l + 1)\sigma^2}{2}\right) \\
    &\quad \times \sin\left(\left(l + \frac{1}{2}\right)\omega\right) \sin\left(\frac{\omega}{2}\right)
    \end{aligned} \\
\label{eq:score_diffkernel}
    \nabla \ln p_t(R' | R) = \left( \frac{d}{d\omega} \log f(\omega) \right) \hat{\omega}
\end{gather}

% \begin{figure*}[t]
%     \centering
%     \includegraphics[width=\textwidth]{figures/results_figure1.pdf}
%     \caption{}
%     \label{fig:adsorbdiff_results}
% \end{figure*}

\subsection{Conditional denoising diffusion for adsorbate placement}
\label{sec:cond_diff}

While the OC Challenge set provides densely calculated adsorption energies for 244 systems, a total of 244 * 100 DFT optimization benchmarks were conducted. This involved performing 100 different random placements for each configuration. Notably, the naive denoising diffusion setup was exclusively trained on the 244 lowest energy configurations.

To leverage the entirety of the DFT optimization data, a conditional diffusion model is employed. In this model, the optimized position is conditioned on the relative energy, specifically relative to the energy of the lowest energy configuration ($E_{\text{rel-i}}^c = E_{\text{min}}^c - E_i^c$). This approach allows for a more comprehensive utilization of the available DFT optimization data.

\subsection{Graph Neural Network (GNN) architecture}
The inputs to the ML model are the 3D positions of all input atoms from the adslab configuration and their corresponding atomic numbers. The outputs predict per-atom 3D vectors. These vectors are forces in the case of force fields and the score function in the case of diffusion. To predict multiple score functions (for translation and rotation), multiple output heads are trained each predicting independent score functions. 

All architectures used in this work come under the message-passing neural network (MPNN) framework of graph neural networks (GNNs). MPNNs operate by passing messages between nodes in the graph, allowing information to be exchanged and aggregated iteratively. The key components of an MPNN include message passing, updating node states, and global readout. In the message-passing step, nodes exchange information based on their local context, and this information is then used to update the states of the nodes (as shown in Equation \ref{eq:mpnn}).

\begin{equation}
\label{eq:mpnn}
h_v^{(t+1)} = \text{Update}\left(h_v^{(t)}, \, \text{Aggregate}\left(\{m_{u \to v}^{(t)} \,|\, u \in \mathcal{N}(v)\}\right)\right)
\end{equation}

Here, \(h_v^{(t)}\) represents embeddings of node \(v\) at iteration \(t\), \(m_{u \to v}^{(t)}\) denotes the message from node \(u\) to \(v\) at iteration \(t\), \(\mathcal{N}(v)\) represents the neighborhood of node \(v\), and \(\text{Update}\) and \(\text{Aggregate}\) are differentiable functions for updating node states and aggregating messages, respectively.

In our study, we systematically investigate diverse architectures employed in the training of diffusion models to discern the significance of architectural decisions in this context. Specifically, we have chosen to assess the performance of PaiNN, GemNet-OC, and EquiformerV2, each distinguished by its treatment of explicit geometric information and rotational symmetries \cite{duval2023hitchhiker}. This selection is grounded in the diverse characteristics they bring to the table. Furthermore, we employ these architectures in benchmarking against OC20 force-field evaluation, thereby facilitating comparative analysis of architectural significance in the realms of force-fields and diffusion.

\section{Results}

In this section, we present results demonstrating the impact of AdsorbDiff in accelerating the search for adsorption energy or better global optima. Specifically, we demonstrate the impact of conditional denoising training over unconditional training and a randomly placed adsorbate baseline. This random baseline is equivalent to performing AdsorbML on a single site (Nsite=1). Additionally, we demonstrate the impact of pretraining, model architectures, and the generalization of this approach to new adsorbates and slabs.
\subsection{Datasets}

We utilize two publicly available datasets for this work - OC20-Dense \cite{lan2023adsorbml} and OC20 \cite{chanussot2021open}. 

\textbf{OC20:} Open Catalyst 2020 (OC20) is a large-scale dataset that contains converged DFT optimization trajectories of 460k unique adslab configurations, encompassing 55 unique elements and 74 adsorbates. Note that these optimizations are local optimizations performed with a single heuristic placement. ML force field models are trained on the forces derived from these DFT trajectories. Additionally, the optimized structure from OC20 is utilized for pre-training the diffusion model.

\textbf{OC20-Dense:} The OC20-Dense dataset serves as a DFT benchmark for adsorption energies, employing dense placement on 100 random sites per adslab configuration, followed by DFT optimization. This dataset releases both in-distribution (ID) and out-of-distribution (OOD) data, relative to OC20. The ID data incorporates adsorbates and slabs from OC20's training set but presents different combinations and configurations, while OOD introduces new adsorbates and/or slabs not found in the OC20 training set. A subset of OC20-Dense ID and OOD was utilized in the Open Catalyst Challenge 2023, hosted at the AI for Science Workshop during NeurIPS 2023 \footnote{\url{https://opencatalystproject.org/challenge.html}}. We split the ID data into 80/20 ratios for training the diffusion model and validating the sampling process. These smaller subsets make it computationally cheaper to perform end-to-end iterations.

\subsection{Metric and constraints}

Our success metric is defined by the final energy calculated through DFT.  For real-world applications, this energy ($D_{Total}^{DFT}$) is used in calculating the adsorption energy $E_{Ads}^{DFT}$ as $E_{Adsorption}^{DFT} = E_{Total}^{DFT} - E_{Slab}^{DFT} - E_{Adsorbate}^{DFT}$, where $E_{Slab}^{DFT}$ and $E_{Adsorbate}^{DFT}$ are the independent energies of slab and adsorbate respectively. This adsorption energy acts as a thermodynamic description of how good a catalyst is for downstream application. The DFT Success Rate (SR) is defined as the percentage of \textit{valid} structures within 0.1 eV or lower of the DFT computed adsorption energy benchmark in the OC20-Dense data (as described in AdsorbML). This is computationally expensive to calculate but is accurate. Metrics calculated from ML predictions are inexpensive but are also inaccurate, discussed further in Appendix \ref{app:metrics}.

Since we calculate adsorption energies, the adsorbate and slab must not change during optimization. Therefore, the structures are considered an \textit{anomaly} due to - (1) adsorbate desorption: adsorbate moves far away from the slab, (2) adsorbate dissociation: atoms in adsorbate dissociate into multiple adsorbates, (3) slab mismatch/reconstruction: slab reconstructs into a completely different structure during optimization (4) adsorbate intercalation: when any of the adsorbate atoms detaches and get into the slab. 

\textbf{Experimental setup:} All presented results are based on the DFT success rate metric as defined in the preceding section. Throughout the diffusion process, we employ the EquiformerV2 architecture, unless explicitly stated otherwise, owing to its state-of-the-art performance in AdsorbML. Additionally, for MLFF optimization, we utilize GemNet-OC pre-trained on OC20, chosen for its lower inference cost. Further specifics regarding model and training hyperparameters are available in Appendix \ref{app:hyperparameter}. All results are shown on the val ID split apart from the OOD section.

\subsection{Conditional vs Unconditional diffusion}
\begin{figure}[h]
    \centering
    \includegraphics[width=\textwidth/2]{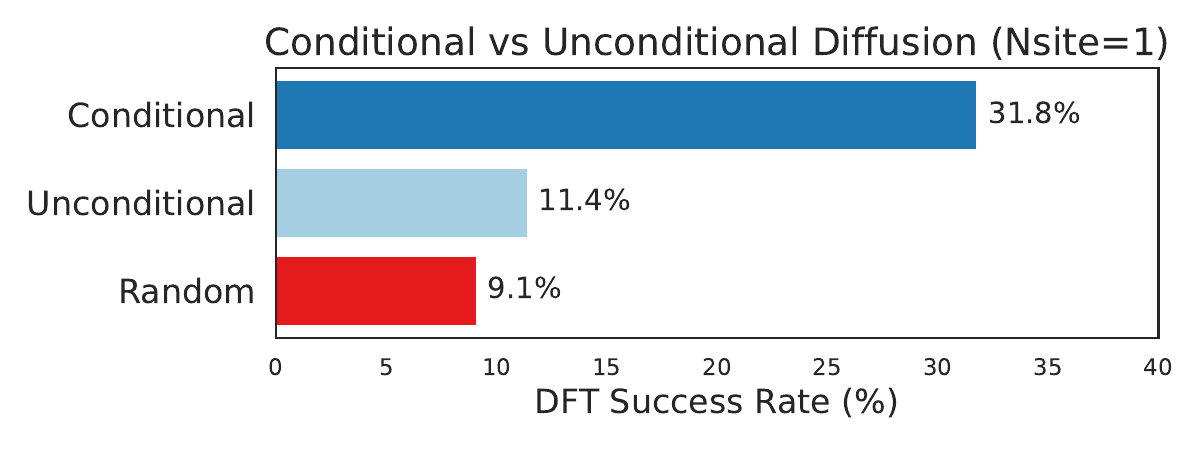}
    \caption{
Comparison of conditional and unconditional diffusion with a baseline of random placement. Conditional diffusion training on relative energies of configurations of adslab significantly improves success rates over unconditional training and AdsorbML baseline.}
    \label{fig:conditional_vs_unconditional}
\end{figure}
We demonstrate the importance of conditional training on relative energies (as shown in Section \ref{sec:cond_diff}) over unconditional diffusion training in Figure \ref{fig:conditional_vs_unconditional}. We compare both of these approaches to a naive baseline of AdsorbML with a single site (Nsite=1) where MLFF optimization is performed on a random adsorbate placement.  It is noteworthy that the performance of unconditional training is suboptimal, this may be ascribed to the unexploited potential of additional data made available through conditional training.

\subsection{AdsorbDiff vs AdsorbML}
\begin{figure}[h]
    \centering
    \includegraphics[width=\textwidth/2]{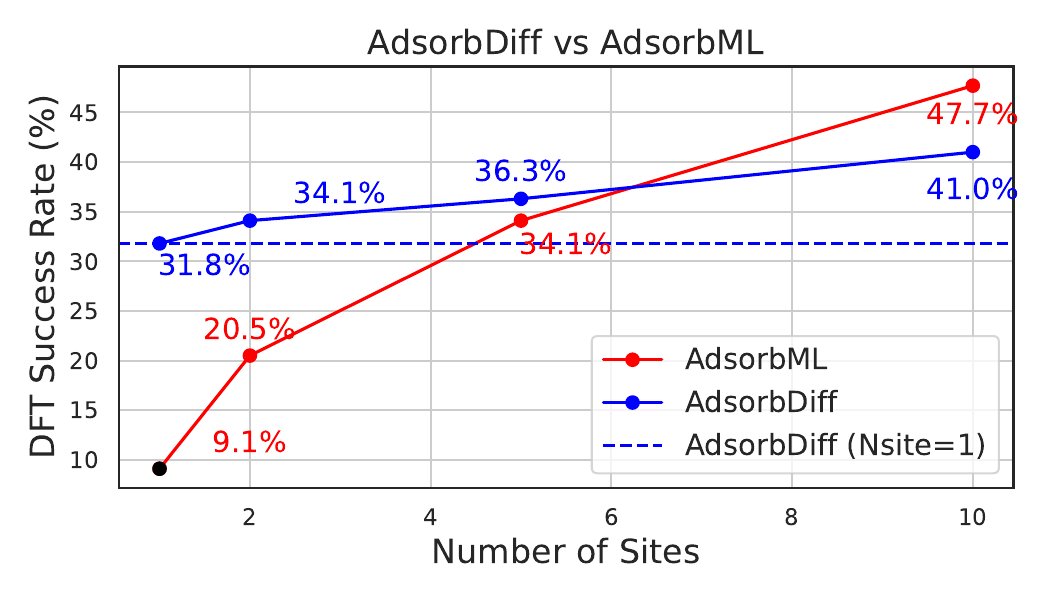}
    \caption{DFT Success Rates (\%) for AdsorbDiff and AdsorbML across a varying number of site predictions. AdsorbDiff performs 3.5x better than AdsorbML utilizing a single site prediction. At higher sites, AdsorbML performs better due to the brute-force nature of site prediction that reduces anomalies.}
    \label{fig:adsorbdiff_vs_adsorbml}
\end{figure}
\begin{figure}[h]
    \centering
    \includegraphics[width=\textwidth/2]{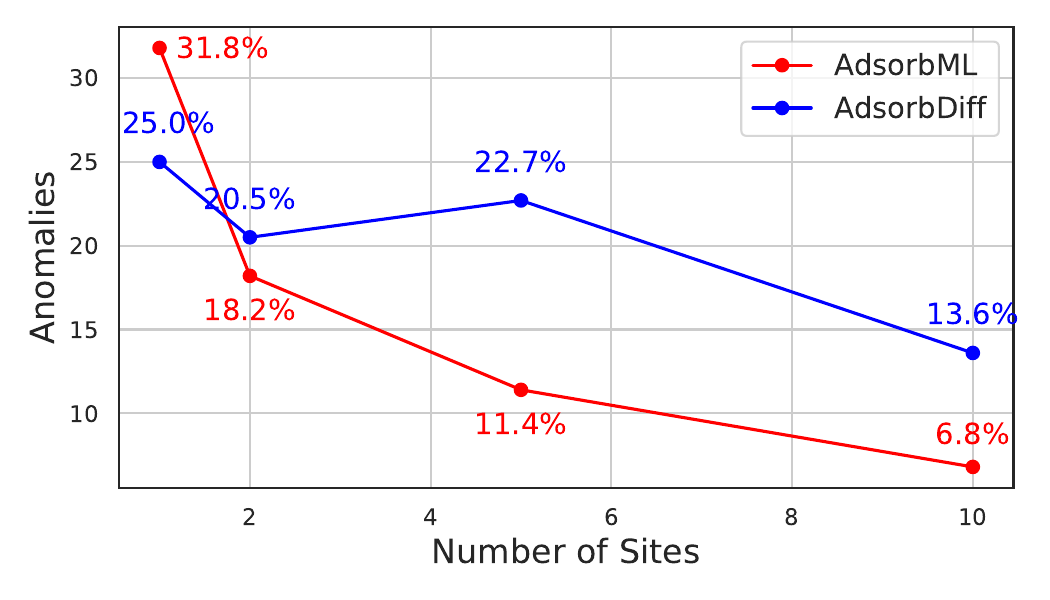}
    \caption{Anomalies in AdsorbDiff and AdsorbML with respect to Nsites. A system is labeled as anomalous if all its predicted sites result in anomalies. AdsorbML has fewer anomalies than AdsorbDiff at higher Nsites due to more randomness in initial sites.}
    \label{fig:anomalies}
\end{figure}

AdsorbML conducts MLFF optimization and DFT evaluations on adsorption sites randomly placed within the system. A comparative analysis is drawn with AdsorbDiff, where the prediction of adsorption sites is facilitated through the utilization of diffusion models. As depicted in Figure \ref{fig:adsorbdiff_vs_adsorbml}, it is evident that AdsorbDiff exhibits notably superior performance, particularly at lower Nsites. However, as the number of adsorption sites (Nsites) increases, AdsorbDiff tends to either converge to or underperform in comparison to the brute force approach employed by AdsorbML. Adsorbate sites sampled from AdsorbDiff have less diversity by design as it's trained to predict the global optima. We calculate the average across the standard deviation of the points sampled at 10 Nsites and get \textbf{8.1~\AA}\ for AdsorbML and \textbf{2.7~\AA}\ for AdsorbDiff. AdsorbML's brute force placements have more randomness which leads to fewer anomalies post the MLFF optimization process shown in Figure \ref{fig:anomalies}.

\subsection{Impact of pretraining}

Conditional diffusion benefits from training on a dataset that is 100 times more extensive than the unconditional approach, a consequence of leveraging multiple local optima within a unique adslab configuration. The substantial increase in training data size manifests in a notable enhancement in the success rate for the conditional approach. The OC20 IS2RE dataset, containing optimization data for 460,000 distinct adslab combinations, serves as a valuable resource for pretraining the diffusion model. It is important to acknowledge that this pretraining process results in a model that learns the local optima of an adslab combination, with the caveat that the model may not capture global optima for an adslab combination.

\begin{figure}[h]
    \centering
    \includegraphics[width=\textwidth/2]{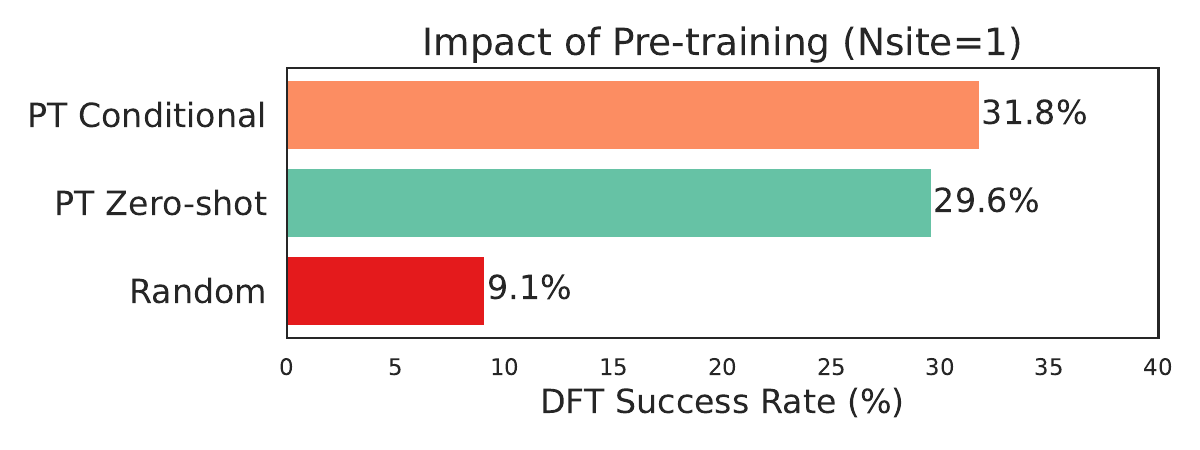}
    \caption{Impact of pretraining on 460k OC20 local optima data on DFT Success Rate. PT Zero-shot measures zero-shot generalization of OC20 pre-trained model to OC20-Dense data. PT Conditional is finetuned on OC20 Dense data conditionally on relative energies of adslab configurations. Random baseline corresponds to randomly placed adsorbate.}
    \label{fig:pretraining_impact}
\end{figure}

% We pre-train a diffusion model utilizing the OC20 IS2RE dataset (as described in Section \ref{sec:diff_method}). Following this, we evaluate the model's impact on downstream diffusion tasks using the OC20-Dense dataset.

\textbf{IS2RS Pretraining (PT) Zero-shot:}
Taking advantage of the diffusion model pre-trained on OC20 IS2RE data, we conduct a zero-shot validation on the OC20-Dense ID val split. This experimental setup allows us to assess the model's ability to predict better global optima having trained on a large dataset of local optima. Notably, we observe a substantial increase in DFT success rate in the zero-shot setting (as shown in Figure \ref{fig:pretraining_impact}).

\textbf{IS2RS Pretraining (PT) Conditional:}
In this approach, we utilize the pre-trained model using the OC20-Dense data as described in Section \ref{sec:cond_diff}. We observe that although this gives a 2\% improvement over zero-shot, it converges to the same results as just training conditionally on OC20-Dense (shown in Figure \ref{fig:pretraining_impact}).

\subsection{Impact of architectures}
\label{sec:arch}

Architectures characterized by richer geometric information and extensive many-body interaction capabilities, such as eSCN and EquiformerV2, have demonstrated superior performance in force evaluations within the OC20 dataset compared to simpler models like PaiNN, which primarily encode directional information and apply linear transformations. Our benchmarking involves the evaluation of three architectures that exhibit progressively improved performance in OC20 Force MAE, revealing significant differences among them.

This evaluation is specifically conducted in the context of the zero-shot assessment following pretraining (PT zero-shot) on an extensive dataset encompassing 460,000 OC20 instances. This choice is inspired by insights from the GemNet-OC paper \cite{gasteiger2022gemnet}, suggesting that certain architectural choices manifest optimal performance only at higher data scales.

\begin{figure}[h]
    \centering
    \includegraphics[width=\textwidth/2]{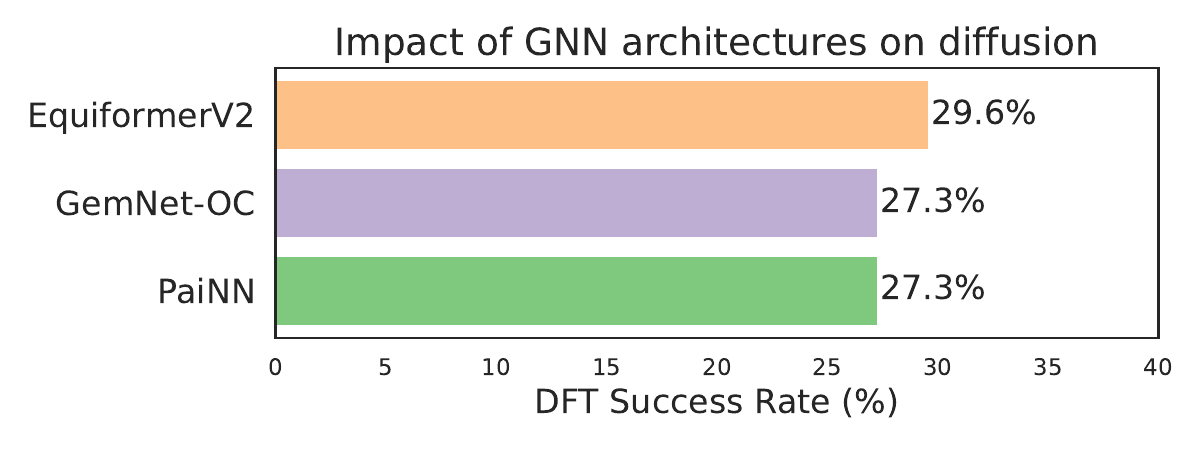}
    \caption{Impact of Graph Neural Network (GNN) architectures on the diffusion process for DFT Success Rate keeping other parts of the framework same. Different architectures perform similarly on the task of diffusion sampling.}
    \label{fig:architecture_diff}
\end{figure}

Interestingly, in the realm of the diffusion task, we note that the disparity in success rates among these architectures is marginal (as shown in Figure \ref{fig:architecture_diff}) which has been recently demonstrated in applications of molecular generation tasks as well \cite{wang2023generating}. The intuition behind this result is that the diffusion model's score function can be thought of as learning a harmonic potential \cite{xie2021crystal}. Harmonic potentials are simpler force-fields than ab-initio DFT calculations involved in OC20 forces. This could result in simpler architectures being able to capture the underlying complexity of the diffusion task defined in our work.

\subsection{OOD generalization}

We measure the success of AdsorbDiff in out-of-distribution (OOD) cases where the model hasn't seen the adsorbate or the slab even during the pre-training on OC20. We pick a random 50 samples out of 200 validation OOD split defined in Open Catalyst Challenge 2023. We observe a marginal decrease of only 3.8\% in results for the OOD case compared to the ID scenario and consistently observe significant improvement over the AdsorbML (Nsite=1) baseline.

\begin{figure}[h]
    \centering
    \includegraphics[width=\textwidth/2]{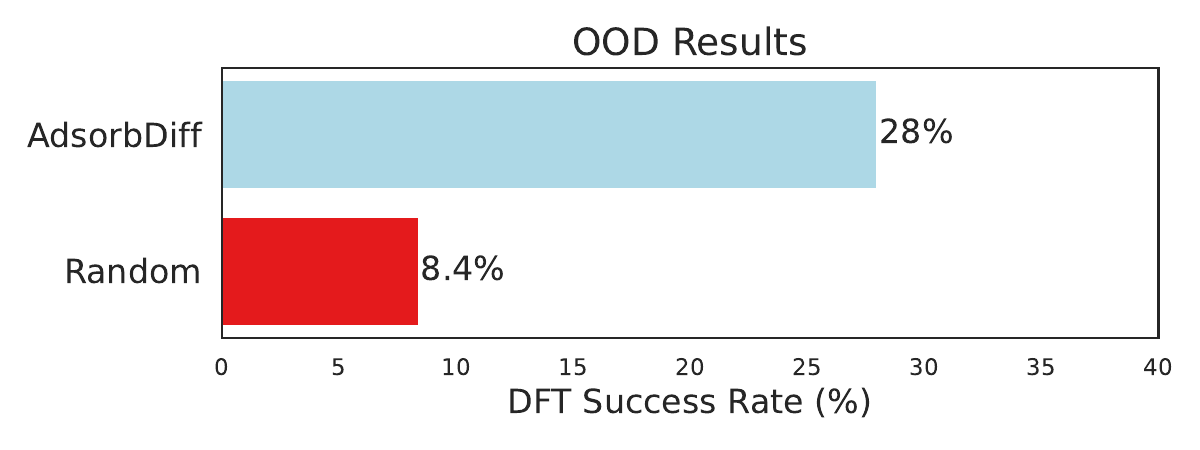}
    \caption{Comparison of DFT Success Rate for In-Distribution (ID) and Out-of-Distribution (OOD) splits using the AdsorbDiff method. Random baseline corresponds to randomly placed adsorbate.}
    \label{fig:ood}
\end{figure}

\subsection{Inference cost}

In the case of conditional diffusion, our approach maintains a maximum step limit of 100, with adsorbate placement converging, on average, within 98 steps. In contrast, for MLFF optimization with a maximum step limit of 300 and Fmax criteria of 0.01 eV/A (consistent with AdsorbML), the convergence occurs in approximately 286 steps. Consequently, for scenarios with a single adsorption site (Nsite 1), AdsorbDiff incurs approximately 34\% more inference cost than AdsorbML, given the GNN architecture for diffusion and MLFF optimization is the same. This end-to-end ML framework is $O(10^4)$ times faster than the conventional DFT pipelines \cite{lan2023adsorbml}.

In Section \ref{sec:arch}, we illustrate that simpler and faster models such as PaiNN yield comparable performance to more intricate and slower models like EquiformerV2. This enhances the efficiency of our diffusion-based approach, as its computational burden becomes negligible in comparison to MLFF optimization, which would require more computationally intensive ML architectures (details in Appendix \ref{app:compute}).

\section{Conclusion}

This work introduces AdsorbDiff, a novel conditional denoising diffusion framework adept at leveraging inherent symmetries in adsorbate and slab interactions, enabling efficient prediction of the lowest energy site. The proposed end-to-end evaluation framework, coupled with Density Functional Theory (DFT), provides a robust assessment of our approach's capability to predict optimal adsorption energies. Notably, AdsorbDiff achieves a remarkable 31.8\% success rate with a single site prediction, surpassing the naive AdsorbML baseline (9.1\%) by 3.5x. We demonstrate the benefits of pretraining on large-scale local optima of adsorption sites. Interestingly, we find the diffusion method's performance to be not significantly dependent on the GNN architecture choice. Furthermore, our model's demonstrated generalization to previously unseen adsorbates and slabs underscores its adaptability and robustness.

\section{Limitations and Future Work}
Our findings emphasize that anomalies play a substantial role in diminishing success rates, particularly in the context of multiple site predictions. While certain works have successfully employed constraints, such as Hookean constraints, to mitigate these anomalies, their implementation in a computationally efficient manner for larger adsorbates remains non-trivial. Addressing this challenge stands out as a crucial avenue for future research. Furthermore, the incorporation of torsion angles presents a promising direction for further improvement, especially when dealing with larger adsorbates. 

% Exploring and refining these approaches holds significant potential for enhancing the overall predictive performance of models in adsorption energy 

\section*{Impact statement}
This work's goal is to accelerate catalyst discovery using machine learning. AdsorbDiff substantially accelerates catalyst search which has a positive impact in the field of developing renewable energy technologies and various chemicals. However, there's a possibility of utilizing this work to accelerate the search for catalysts for hazardous chemicals. 

\section*{Acknowledgements}

We thank Minkai Xu, Muhammed Shuaibi, Nima Shoghi, Abhishek Das, and the FAIR Chemistry team at Meta for their valuable feedback and discussions. 

\bibliography{main}
\bibliographystyle{icml2024}

%%%%%%%%%%%%%%%%%%%%%%%%%%%%%%%%%%%%%%%%%%%%%%%%%%%%%%%%%%%%%%%%%%%%%%%%%%%%%%%
%%%%%%%%%%%%%%%%%%%%%%%%%%%%%%%%%%%%%%%%%%%%%%%%%%%%%%%%%%%%%%%%%%%%%%%%%%%%%%%
% APPENDIX
%%%%%%%%%%%%%%%%%%%%%%%%%%%%%%%%%%%%%%%%%%%%%%%%%%%%%%%%%%%%%%%%%%%%%%%%%%%%%%%
%%%%%%%%%%%%%%%%%%%%%%%%%%%%%%%%%%%%%%%%%%%%%%%%%%%%%%%%%%%%%%%%%%%%%%%%%%%%%%%

\appendix
\onecolumn
\section*{Table of Contents}
\begin{itemize}
    \item[] \ref{app:background}\quad Background in catalysis
    \item[] \ref{app:compute}\quad Compute details 
    \item[] \ref{app:metrics}\quad Metrics other than DFT Success Rate
    \item[] \ref{app:hyperparameter}\quad Model and training hyperparameters

\end{itemize}
\section{Background in catalysis}
\label{app:background}

Heterogeneous catalysis is a pivotal process in chemistry and industry. Unlike its counterpart, homogeneous catalysis, where reactants, products, and the catalyst coexist in the same phase, heterogeneous catalysis involves catalysts that exist in a different phase from the reactants or products. This phase distinction extends beyond solid, liquid, and gas components—it also encompasses immiscible mixtures (such as oil and water) or any scenario where an interface is present. In most cases, heterogeneous catalysis involves solid-phase catalysts and gas-phase reactants. The heart of this process lies in a cycle of molecular adsorption, reaction, and desorption occurring at the catalyst surface. Thermodynamics, mass transfer, and heat transfer all influence the rate of these reactions. Heterogeneous catalysis plays a crucial role in large-scale production and selective product formation, impacting approximately 35\% of the world’s GDP \cite{ma2006heterogeneous} and aiding in the production of 90\% of chemicals by volume \cite{rothenberg2017catalysis}.

\begin{figure}[h]
    \centering
    \includegraphics[width=0.8\textwidth]{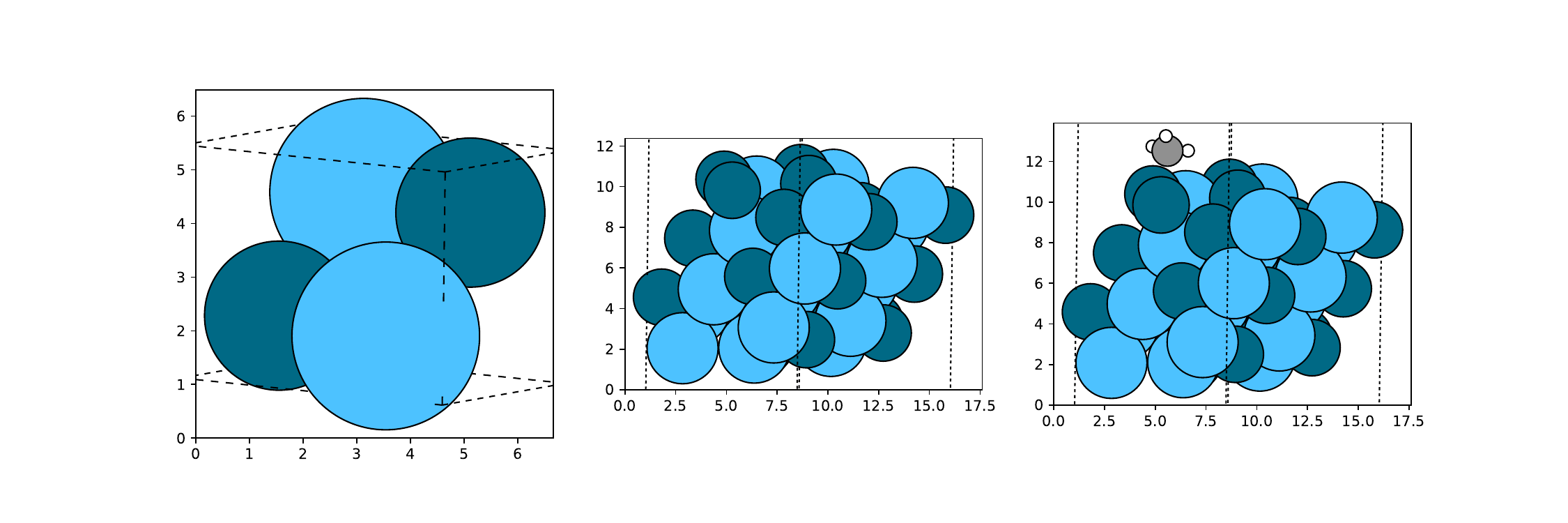}
    \caption{This is a system sampled from the OC20-Dense data. Left: An example bulk structure of Hf2Pd2. Center: Slab created out of the bulk. Right: CH3 adsorbate randomly placed on the slab.}
    \label{fig:ood}
\end{figure}
\subsection{Bulks}
The bulk refers to the solid material that serves as the foundation for catalytic reactions. It constitutes the underlying structure upon which the catalyst operates. Think of it as the bedrock—the stable, three-dimensional lattice of atoms or molecules that provides the framework for further interactions.

\subsection{Slabs}
A slab is a specific surface layer of the bulk material. Imagine slicing through the bulk crystal lattice to reveal a flat, exposed surface. Slabs are essential because they expose a large number of active sites where reactants can interact with the catalyst. These surfaces play a critical role in determining the overall catalytic behavior. Researchers often explore different slab orientations (such as (100), (111), or (110)) to understand how the arrangement of atoms affects reactivity.

\subsection{Adsorbates}
An adsorbate is a molecule or atom that adheres to the surface of the catalyst. During heterogeneous catalysis, reactant molecules (the adsorbates) bind to specific sites on the catalyst surface (the adsorbent). This binding occurs via van der Waals forces, including dipole-dipole interactions, induced dipole interactions, and London dispersion forces. Importantly, no chemical bonds form between the adsorbate and adsorbent; their electronic states remain relatively unperturbed. Adsorption is a crucial step in the catalytic process, as it positions reactants near active sites for subsequent reactions.

\section{Compute details and discussions}
\label{app:compute}

Do AdsorbDiff speedups impact practically?

Supplementary Table VII of AdsorbML \cite{lan2023adsorbml} discusses the computational cost of MLFF optimization (or ML RX as mentioned in that work) and DFT SP / RX costs for different GNN architectures. The computational cost for 300 steps across 100 Nsites of ML RX of PaiNN is 60.4 GPU-hours on 32GB NVIDIA V100 cards. The cost for the best models SCN-MD-Large and GemNet-OC-MD-Large is 1129.2 and 638.3 respectively. The cost of performing diffusion with PaiNN at 1 site (60.4 / 300 GPU-hours) is insignificant to the MLFF optimization cost at 1 Nsite of any of the recent models like SCN-MD-Large (1129.2 / 100 GPU-hours). 

Why should we aim to perform only a few MLFF optimizations?

With the MLFF models getting larger, more complex, and more accurate, their inference costs have also increased. The assumption that MLFF optimization inference costs are completely insignificant in comparison to DFT is not valid for larger models. If we assume that only DFT SP is required for the purpose of ranking sites, we can see in the same table of AdsorbML that for SCN-MD-Large the DFT-SP cost is 2645.29 CPU-hours performed across 5 sites. Given that DFT on GPUs is getting faster and it's roughly 5x faster for these systems, DFT SP verifications (2645.39 / 5 GPU-hours) are not truly the bottleneck in comparison to ML RX (1129.2 GPU-hours). To be able to perform billions of these simulations, it's critical to make the ML simulations significantly more efficient.

This hints at a potential progression in future computational workflows, wherein they would commence with a diffusion-based simulation leveraging simple and fast ML architectures. Following this, MLFF optimization might utilize foundation models targeting atomic structures\cite{shoghi2023molecules, batatia2023foundation}, culminating in the validation of results through a select number of DFT calculations.
\section{Metrics other than DFT Success Rate}
\label{app:metrics}
We present all our findings using the DFT Success Rate within our end-to-end framework, recognizing the significance of DFT metrics over purely ML-based metrics, which may suffer from inaccuracies. In this section, we offer the ML Success Rate for key results, comparing it with the DFT Success Rate. Two metrics, solely tracked with our ML models without DFT, are considered. Firstly, the success metric based on the distance between the optimal site and predicted site through diffusion. We define a prediction as successful if the Mean Absolute Error (MAE) distance is less than 1 \AA. Secondly, we assess it after MLFF optimization using ML predicted energies. Note that the MLFF optimization is performed using GemNet-OC checkpoint pretraining on OC20 2M splits across all experiments for consistency.

\begin{table}[h]
    \centering
    \begin{tabular}{|l|c|c|c|}
        \hline
        \textbf{Method} & \textbf{Diff SR} & \textbf{MLFF SR} & \textbf{DFT SR} \\
        \hline
        \hline
        Unconditional & 15.9 & 22.72 & 11.36 \\
        Conditional & 36.4 & 29.5 & 31.81 \\
        PT zero-shot & 38.63 & 27.27 & 29.55 \\
        PT conditional & 29.5 & 47.72 & 31.81 \\
        \hline
        PaiNN & 29.5 & 27.3 & 27.3 \\
        GemNet & 29.5 & 27.3 & 27.3 \\
        Equiformer V2 & 29.5 & 27.3 & 29.6 \\
        \hline
    \end{tabular}
    \caption{Comparison of key results with different metrics for a single-site prediction}
    \label{tab:metrics}
\end{table}

We also perform an experiments where we start from the optimal site and perform MLFF optimization and calculate its MLFF Success Rate and find it to be \textbf{43\%}. With DFT, its 100\%. Therefore, we discourage the use of only ML based metrics in evaluating these results. This also demonstrates the importance of our end-to-end framework.
\section{Model and training hyperparameters}
\label{app:hyperparameter}

In the case of diffusion training, an additional head is incorporated for each of these architectures to account for the two vector predictions needed – one for the translation vector and the other for the Euler vector.

\subsection{EquiformerV2}
Definitions of the attributes mentioned in Table \ref{tab:equiformer_hyperparams} can be found in the open-sourced code of the original work \url{https://github.com/Open-Catalyst-Project/ocp/tree/main/ocpmodels/models/equiformer_v2}.

\begin{table}[h]
    \centering
    \begin{tabular}{|l|c|}
        \hline
        \textbf{Hyperparameter} & \textbf{Value} \\
        \hline
        \hline
        \texttt{max\_neighbors} & 20 \\
        \texttt{max\_radius} & 12.0 \\
        \texttt{max\_num\_elements} & 90 \\
        \hline
        \texttt{num\_layers} & 8 \\
        \texttt{sphere\_channels} & 128 \\
        \texttt{attn\_hidden\_channels} & 64 \\
        \texttt{num\_heads} & 8 \\
        \texttt{attn\_alpha\_channels} & 64 \\
        \texttt{attn\_value\_channels} & 16 \\
        \texttt{ffn\_hidden\_channels} & 128 \\
        \texttt{norm\_type} & 'layer\_norm\_sh' \\
        \hline
        \texttt{lmax\_list} & [4] \\
        \texttt{mmax\_list} & [2] \\
        \texttt{grid\_resolution} & 18 \\
        \hline
        \texttt{num\_sphere\_samples} & 128 \\
        \hline
        \texttt{edge\_channels} & 128 \\
        \texttt{use\_atom\_edge\_embedding} & True \\
        \texttt{distance\_function} & 'gaussian' \\
        \texttt{num\_distance\_basis} & 512 \\
        \hline
        \texttt{attn\_activation} & 'silu' \\
        \texttt{use\_s2\_act\_attn} & False \\
        \texttt{ffn\_activation} & 'silu' \\
        \texttt{use\_gate\_act} & False \\
        \texttt{use\_grid\_mlp} & True \\
        \hline
        \texttt{alpha\_drop} & 0.1 \\
        \texttt{drop\_path\_rate} & 0.1 \\
        \texttt{proj\_drop} & 0.0 \\
        \hline
        \texttt{weight\_init} & 'uniform' \\
        \hline
    \end{tabular}
    \caption{EquiformerV2 hyperparameters}
    \label{tab:equiformer_hyperparams}
\end{table}

\subsection{GemNet-OC}
Definitions of the attributes mentioned in Table \ref{tab:gemnet_hyperparams} can be found in the open-sourced code of the original work \url{https://github.com/Open-Catalyst-Project/ocp/tree/main/ocpmodels/models/gemnet_oc}. The same model architecture hyperparameters are utilized for diffusion and MLFF optimization.

\subsection{PaiNN}
Definitions of the attributes mentioned in Table \ref{tab:painn_hyperparams} can be found in the open-sourced code of the work adapted in the OCP repository \url{https://github.com/Open-Catalyst-Project/ocp/tree/main/ocpmodels/models/painn}.

\begin{table}[h]
    \centering
    \begin{tabular}{|l|c|}
        \hline
        \textbf{Hyperparameter} & \textbf{Value} \\
        \hline
        \hline
        \texttt{hidden\_channels} & 512 \\
        \texttt{num\_layers} & 6 \\
        \texttt{num\_rbf} & 128 \\
        \texttt{cutoff} & 12.0 \\
        \texttt{max\_neighbors} & 50 \\
        \hline
    \end{tabular}
    \caption{PaiNN hyperparameters}
    \label{tab:painn_hyperparams}
\end{table}

\subsection{MLFF optimization}

All MLFF optimization in this study is conducted using the pre-trained GemNet-OC model, trained on the OC20 2M dataset. The preference for GemNet-OC over the more recent EquiformerV2 architecture stems from its up to 8x faster inference speeds, coupled with reasonably good accuracies. This choice shouldn't impact the qualitative results of this work as this method is model agnostic. 

Optimizations are performed using L-BFGS algorithm \cite{liu1989limited}. The hyperparameters of L-BFGS are mentioned in Table \ref{tab:lbfgs}. During MLFF optimization, we move both the surface atoms and adsorbate atoms. We keep a maximum step of 300 and an F max of 0.01 eV/\AA. All of these settings are consistent across all our runs.

\begin{table}[h]
    \centering
    \begin{tabular}{|l|c|}
        \hline
        \textbf{Hyperparameter} & \textbf{Value} \\
        \hline
        \hline
        \texttt{maxstep} & 0.04 \\
        \texttt{memory} & 50 \\
        \texttt{damping} & 1.0 \\
        \texttt{alpha} & 70.0 \\
        \hline
    \end{tabular}
    \caption{L-BFGS hyperparameters}
    \label{tab:lbfgs}
\end{table}

\subsection{Sampling hyperparameters}

We sample from the SDE described in Section \ref{sec:diff_method}. In practice, we find the ODE formulation to be more performant across all experiments and therefore that's our default sampling approach. This could be due to the global optima search that we aim for instead of just sampling from the distribution of minima. We set a max step size for sampling and stop early if the movement of atoms converges. These hyperparameters are shown in Table \ref{tab:sampling_hyperparams}. 

We have also tried the Annealed Langevin Sampling formulation proposed in Noise Conditioned Score Network (NCSN) \cite{song2019generative} but find it to be suboptimal taking 10x more steps for convergence and doesn't generalize well outside of the domain of training data. 

\begin{table}[h]
    \centering
    \begin{tabular}{|l|c|}
        \hline
        \textbf{Hyperparameter} & \textbf{Value} \\
        \hline
        \hline
        \texttt{Number of sampling steps} & 100 \\
        \texttt{Translation low std} & 0.1 \\
        \texttt{Translation high std} & 10 \\
        \texttt{Rotation low std} & 0.01 \\
        \texttt{Rotation high std} & 1.55 \\
        \hline
    \end{tabular}
    \caption{Sampling hyperparameters}
    \label{tab:sampling_hyperparams}
\end{table}

\subsection{Training compute and hyperparameters}

We perform all of our training on 2 48GB A6000 GPUs. We use cosine learning rate with linear warmup. The conditional diffusion model has been trained for 47 GPU-hours. Maximum learning rate was kept to be 4.e-4 and 1.e-4 for pretraining and finetuning runs respectively utilizing AdamW optimizer. For each model architecture, maximum batch size was utilized that could fit on the GPU. Overall, we keep the training methods similar to the original works of respective model architectures. 

\begin{table}[h]
    \centering
    \begin{tabular}{|l|c|}
        \hline
        \textbf{Hyperparameter} & \textbf{Value} \\
        \hline
        \hline
        \texttt{num\_spherical} & 7 \\
        \texttt{num\_radial} & 128 \\
        \texttt{num\_blocks} & 4 \\
        \texttt{emb\_size\_atom} & 256 \\
        \texttt{emb\_size\_edge} & 512 \\
        \texttt{emb\_size\_trip\_in} & 64 \\
        \texttt{emb\_size\_trip\_out} & 64 \\
        \texttt{emb\_size\_quad\_in} & 32 \\
        \texttt{emb\_size\_quad\_out} & 32 \\
        \texttt{emb\_size\_aint\_in} & 64 \\
        \texttt{emb\_size\_aint\_out} & 64 \\
        \texttt{emb\_size\_rbf} & 16 \\
        \texttt{emb\_size\_cbf} & 16 \\
        \texttt{emb\_size\_sbf} & 32 \\
        \texttt{num\_before\_skip} & 2 \\
        \texttt{num\_after\_skip} & 2 \\
        \texttt{num\_concat} & 1 \\
        \texttt{num\_atom} & 3 \\
        \texttt{num\_output\_afteratom} & 3 \\
        \texttt{cutoff} & 12.0 \\
        \texttt{cutoff\_qint} & 12.0 \\
        \texttt{cutoff\_aeaint} & 12.0 \\
        \texttt{cutoff\_aint} & 12.0 \\
        \texttt{max\_neighbors} & 30 \\
        \texttt{max\_neighbors\_qint} & 8 \\
        \texttt{max\_neighbors\_aeaint} & 20 \\
        \texttt{max\_neighbors\_aint} & 1000 \\
        \texttt{rbf.name} & gaussian \\
        \texttt{envelope.name} & polynomial \\
        \texttt{envelope.exponent} & 5 \\
        \texttt{cbf.name} & spherical\_harmonics \\
        \texttt{sbf.name} & legendre\_outer \\
        \texttt{extensive} & True \\
        \texttt{output\_init} & HeOrthogonal \\
        \texttt{activation} & silu \\
        \hline
        \texttt{regress\_forces} & True \\
        \texttt{direct\_forces} & True \\
        \texttt{forces\_coupled} & False \\
        \hline
        \texttt{quad\_interaction} & True \\
        \texttt{atom\_edge\_interaction} & True \\
        \texttt{edge\_atom\_interaction} & True \\
        \texttt{atom\_interaction} & True \\
        \hline
        \texttt{num\_atom\_emb\_layers} & 2 \\
        \texttt{num\_global\_out\_layers} & 2 \\
        \texttt{qint\_tags} & [1, 2] \\
        \hline
    \end{tabular}
    \caption{GemNet-OC hyperparameters}
    \label{tab:gemnet_hyperparams}
\end{table}

% \section{Additional results}
% \subsection{Loss coefficients for conditional diffusion}

% \section{Case studies}
% Cu(100) on O/H
%Cu(111) on O - FCC/HCP
%CuPd(single atom)[111] - site on Pd
% stepped surfaces
% Acknowledge limitations

%%%%%%%%%%%%%%%%%%%%%%%%%%%%%%%%%%%%%%%%%%%%%%%%%%%%%%%%%%%%%%%%%%%%%%%%%%%%%%%
%%%%%%%%%%%%%%%%%%%%%%%%%%%%%%%%%%%%%%%%%%%%%%%%%%%%%%%%%%%%%%%%%%%%%%%%%%%%%%%

\end{document}